\documentclass{article}

\usepackage[preprint]{corl_2026}

\usepackage[utf8]{inputenc} %
\usepackage[T1]{fontenc}    %
\usepackage{hyperref}       %
\usepackage{url}            %
\usepackage{booktabs}       %
\usepackage{amsfonts}       %
\usepackage{nicefrac}       %
\usepackage{microtype}      %
\usepackage[dvipsnames]{xcolor}
\usepackage{mathtools}
\usepackage{graphicx}
\usepackage{amsmath}
\usepackage{cleveref}
\usepackage{bm}
\usepackage{multirow}
\usepackage{makecell}
\usepackage{wrapfig}
\usepackage{csquotes}
\usepackage{adjustbox}
\usepackage[dvipsnames,table]{xcolor}

\DeclareMathOperator*{\argmin}{arg\,min}
\newcommand{\R}[0]{\mathbb{R}}
\newcommand{\B}[0]{\mathbb{B}}

\DeclarePairedDelimiter{\norm}{\lVert}{\rVert}

\definecolor{impcol}{HTML}{1B7A3A}
\definecolor{degcol}{HTML}{B22B2B}
\definecolor{neucol}{HTML}{8C8C8C}
\definecolor{lowprob}{HTML}{00224E}
\newcommand{\dgood}[1]{{\scriptsize\textcolor{impcol}{\,($-$#1\%)}}}
\newcommand{\dbad}[1]{{\scriptsize\textcolor{degcol}{\,($+$#1\%)}}}
\newcommand{\dflat}[1]{{\scriptsize\textcolor{neucol}{\,($\pm$#1\%)}}}

\newcommand{\std}[1]{{\,\tiny$\pm$#1}}

\usepackage{dsfont}
\newcommand{\indicator}[0]{\mathds{1}}

\definecolor{impcol}{HTML}{1B7A3A}
\definecolor{degcol}{HTML}{B22B2B}
\definecolor{neucol}{HTML}{8C8C8C}
\definecolor{lowprob}{HTML}{00224E}

\usepackage{listings}
\definecolor{lightgray}{gray}{0.95}
\usepackage{tcolorbox}
\tcbuselibrary{breakable}
\newtcolorbox{llmprompt}[1]{%
  breakable,
  colback=blue!5, colframe=blue!30, boxrule=1.5pt, width=\textwidth, 
  title=\vspace{3pt}#1, fonttitle=\bfseries,
  arc=2mm,
  fontupper=\small
}
\newtcolorbox{llmresponse}[1]{%
  breakable,
  colback=black!5, colframe=black!30, boxrule=1.5pt, width=\textwidth, 
  title=\vspace{3pt}#1, fonttitle=\bfseries,
  arc=2mm,
  fontupper=\small
}

\title{Driving, Fast or Slow? Neuro-Symbolic Guidance for Motion Prediction in Multi-Modal Ground Mobility}

\author{%
    Simon Kohaut$^{1}$ \quad Felix Divo$^{1}$ \quad Julius Hahnewald$^{1}$ \quad Benedict Flade$^{2}$ \\ 
    \textbf{Julian Eggert}$^{2}$ \quad \textbf{Kristian Kersting}$^{1,3,4,5}$ \quad \textbf{Devendra Singh Dhami}$^{6}$\\
    $^{1}$Artificial Intelligence and Machine Learning Lab, TU Darmstadt \quad
    $^{2}$Honda Research Institute \\ 
    $^{3}$Hessian Center for AI (hessian.AI) \quad $^{4}$Centre for Cognitive Science \\
    $^{5}$German Center for AI (DFKI) \quad
    $^{6}$Uncertainty in Artificial Intelligence Lab, TU Eindhoven\\
}

\begin{document}

\maketitle

\begin{abstract}
Accurate and interpretable motion prediction for heterogeneous traffic spaces, including pedestrians, bicycles, cars, and trucks, is essential for safe autonomous navigation.
Nevertheless, state-of-the-art approaches remain predominantly black-box, lacking explicit encoding of the regulatory and behavioral constraints of real-world mobility. 
We propose Trajectory Compliance-Shaping (TraCS), a neuro-symbolic framework that augments existing black-box motion prediction backbones with interpretable and probabilistic first-order logic.
To do so, TraCS employs an agentic code-generation pipeline to bridge the gap between natural-language descriptions of traffic regulations and probabilistic motion prediction. 
Furthermore, TraCS employs a reactive data-streaming inference engine that maintains and efficiently updates compliance landscapes as scenes evolve. 
To prevent TraCS from overconfidently steering the backbone's predictions in the wrong direction, we propose a neural confidence rating learned as a context-aware attenuation of the compliance signal. 
We demonstrate on the Argoverse 2 benchmark how TraCS consistently improves state-of-the-art prediction backbones, showing that probabilistic and symbolic compliance reasoning is a broadly applicable and computationally efficient complement to purely neural motion predictors.
\end{abstract}

\section{Introduction}

Predicting the future motion of traffic agents is a fundamental challenge in autonomous driving and intelligent transportation systems~\citep{DBLP:conf/cvpr/Zhou0LH23qcnet,DBLP:conf/cvpr/TangKSJB024hpnet,zhang2026polaris}. 
Modern approaches have achieved impressive empirical performance by learning trajectory distributions from large-scale datasets~\citep{Argoverse2,DBLP:conf/iccv/EttingerCCLZPCS21}, yet they remain largely agnostic to the explicit nature of rule-governed traffic environments. 
Agents do not merely follow statistical patterns; instead, they navigate within a web of regulatory constraints, right-of-way conventions, and implicit social contracts that vary by agent class, jurisdiction, and local context~\citep{DBLP:journals/ijon/GomesW25}. 
Traffic environments constitute a highly structured instance of a broader class of social systems in which behavior is governed by shared norms, regulations, and collective expectations. 
Developing computational mechanisms that connect natural-language descriptions of such normative structures to executable reasoning processes, therefore, represents an important step toward socially aware autonomous systems.
This structural gap manifests most acutely in safety-critical scenarios and at longer prediction horizons, where trajectory likelihoods diverge sharply between compliant and non-compliant behavior~\citep{DBLP:journals/mlc/ShiCWSLXW26,DBLP:conf/cvpr/WangZ0X23}.
However, purely neural models lack the machinery to distinguish between them.

More broadly, the challenge of aligning learned behavior with explicit normative knowledge extends beyond motion prediction. 
Future autonomous systems will increasingly operate in environments shaped by interactions between humans and artificial agents, where behavior is constrained by regulations, conventions, and shared expectations. 
We therefore consider traffic as an attractive testbed for investigating how natural-language descriptions of norms can be translated into executable reasoning mechanisms that influence agent behavior.

We address this gap via Trajectory Compliance-Shaping (TraCS), a neuro-symbolic augmentation to state-of-the-art motion forecasting models.
To this end, TraCS is designed as a plug-and-play module that integrates natural-language prompts about local traffic regulations to align a backbone's weight distribution across its proposed trajectories with agreed-upon traffic conventions. 
To do so, TraCS leverages an agentic coding pipeline to generate declarative first-order logic programs that describe the compliant behavior of each mobility mode in a given traffic context. 
Rather than operating over raw observations, these programs are grounded in hybrid probabilistic atoms obtained by fitting statistical models to scene features, yielding a principled probabilistic logic layer that captures the uncertainty inherent in real-world environments. 
In addition to bridging the gap between natural language rule descriptions and black-box prediction models, TraCS increases trust by providing inspectable intermediate representations~\citep{DBLP:journals/corr/abs-2412-05152}.

Through a learned neural confidence rating, TraCS's compliance density further governs how strongly TraCS modulates the backbone's trajectory distribution, ensuring  graceful degradation in the face of imperfect encodings, non-compliant traffic participants, or wrongful traffic rule prompts.
In summary, we make the following contributions:
1. We propose Trajectory Compliance-Shaping~(TraCS), a Neuro-Symbolic plug-and-play guidance system that enriches state-of-the-art motion prediction backbones with a symbolic and probabilistic reasoning apparatus. 2. With TraCS, we integrate hybrid probabilistic spatial relations over observed and predicted states and state transitions as a robust reasoning vocabulary for expressing traffic regulations in first-order logic programs. 3. We show how TraCS employs state-of-the-art large language models~(LLMs) and reactive reasoning schemes~\citep{DBLP:journals/corr/kohautReactive} for effective interfacing and online reasoning, respectively, by probabilistically reweighting the backbone's predictions.
4. Finally, we provide a complete reference implementation of TraCS\footnote{\url{https://www.github.com/simon-kohaut/TraCS}}.

\section{Related Work}

\paragraph{Motion Prediction in Ground Mobility.}
Although probabilistic filtering and forecasting~\citep{Thrun2005,kalman1960new,Davison2007,Martinez2005,del1997nonlinear} has been successfully extended to multi-target tracking~\citep{DBLP:journals/tsp/VoM06gmmPHDfilter}, semantic associations~\citep{DBLP:journals/ras/ChenXD22semantifPHDfilter}, and logical constraints~\citep{DBLP:conf/iros/NittiLR13hybridFilter,DBLP:conf/iros/KohautDFDEK25cofi}, recent works have shifted their focus towards learned process models.
As such, forecasting models based on Transformer~\citep{DBLP:conf/nips/VaswaniSPUJGKP17attention} and Mamba~\citep{DBLP:journals/corr/gu2024mamba}, such as QCNet~\citep{DBLP:conf/cvpr/Zhou0LH23qcnet}, HPNet~\citep{DBLP:conf/cvpr/TangKSJB024hpnet}, DeMo~\citep{DBLP:conf/nips/ZhangSZ24demo}, and Polaris~\citep{zhang2026polaris}, have taken the lead in motion prediction benchmarks such as Argoverse 1~\citep{Argoverse}, 2~\citep{Argoverse2, TrustButVerify}, and the Waymo Motion Prediction Challenge~\citep{DBLP:conf/iccv/EttingerCCLZPCS21,Kan_2024_icra}.
Recently, LLMs have been used to directly predict motion trajectories~\citep{liu2025reasonplan,DBLP:conf/cvpr/BaeLJ24} or city-scale trajectories~\citep{DBLP:journals/corr/abs-2605-10020-TrajDLM,DBLP:journals/corr/abs-2605-10782-TrajPrism} by leveraging their high associative capacity but lacking explicit traffic rules.
While approaches such as NeSyMoF~\citep{DBLP:conf/iros/DoulaYMG24nesymof} have considered extracting a set of symbolic rules on top of the agent and environment encodings, they leave the final weighting of hypotheses to the neural apparatus.
Moreover, such neuro-symbolic forecasters and differentiable rule layers require access to gradients and joint training with the predictor and therefore cannot be applied to frozen third-party backbones.
Instead, TraCS consumes natural-language traffic regulations and reweights the hypotheses of pre-trained predictors.

\paragraph{Neuro-Symbolic Representation and Reasoning.}
Neuro-symbolic systems intertwine symbolic reasoning with the sub-symbolic capabilities of deep learning models.
To embrace uncertainty into programmatic logic, such as Prolog~\citep{colmerauer1990introduction}, systems such as Bayesian Logic Programs~\citep{bayesian_logic} and Probabilistic Logic Programs~\citep{problog,inference_in_plp} have been introduced.
Leveraging neural models in tandem with first-order logic programs has sparked multiple research endeavors in the past.
For instance, neuro-symbolic reasoning systems have expanded on the statistical-relational AI paradigm, leading to fully differentiable reasoning apparatuses such as DeepProbLog~\citep{DBLP:conf/nips/ManhaeveDKDR18deepproblog}, NeurASP~\citep{DBLP:conf/ijcai/YangIL20neurasp}, Scallop~\citep{DBLP:journals/pacmpl/LiHN23scallop}, and SLASH~\citep{DBLP:journals/jair/SkryaginODK23slash}.
While some of them are backed by traditional reasoning systems such as Prolog~\citep{colmerauer1990introduction}, modern systems instead often rely on generative semantics via Answer Set Programming (ASP)~\citep{DBLP:journals/jfp/Shah04asp,DBLP:journals/tplp/GebserKKS19}.
More recently, neural solvers have been shown to accelerate and scale the neuro-symbolic paradigm beyond the problems that CPU-bound systems may handle~\citep{DBLP:journals/corr/skryaginASN}, and reactive inference mechanisms for general probabilistic modeling~\citep{DBLP:journals/jossw/BagaevPV23rxinfer}.
In TraCS, we exploit the reactive paradigm for timely and exact inference over the solution space of ASP~\citep{DBLP:journals/corr/kohautReactive}.

\paragraph{Agentic Program Synthesis.}
Logic languages such as ASP are proven tools when modeling various application domains and their inherent constraints.
While manual coding may be an option when the required expert knowledge and time are available, modern LLM-based agentic settings have been shown to effectively generate such representations~\citep{DBLP:conf/kr/IshayY023generateASP}.
To this end, one may further fine-tune existing LLMs for the ASP generation task~\citep{DBLP:conf/kr/CoppolilloC0PR24finetuneLLMtoASP}, or constrain their generation on a formal syntax description~\citep{DBLP:conf/acl/RabinovichSK17grammarConstrain}.
Nevertheless, one-shot generation is oftentimes not robust enough to form correct programs.
Hence, in TraCS, we adopt an agentic approach to generate first-order logic code to systematically reason about tool-based syntax verification, similar to, e.g., MCP-Solver~\citep{DBLP:conf/sat/Szeider25MCPsolver}, and to provide semantic feedback based on an LLM-judgment~\citep{DBLP:conf/nips/ZhengC00WZL0LXZ23llmJudge} over the generated code.

\begin{figure}[t]
    \centering
    \includegraphics[width=\linewidth]{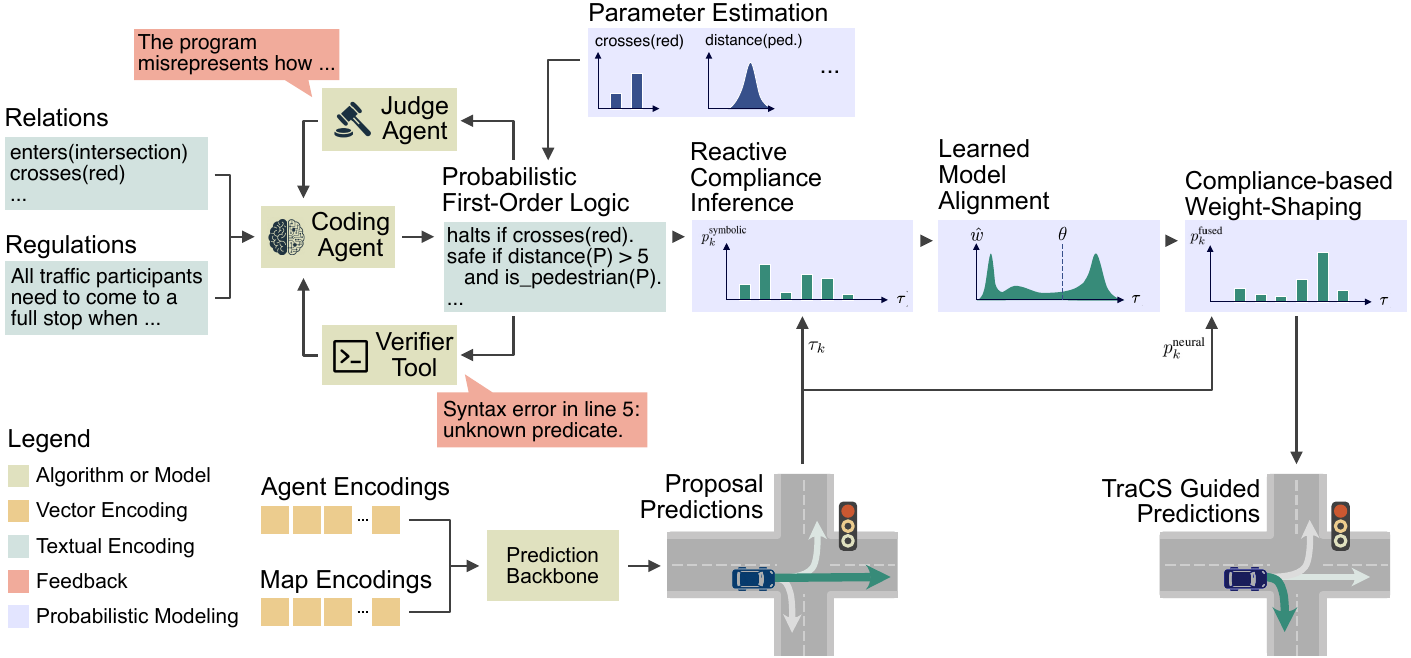}
    \caption{
        \textbf{Probabilistic motion prediction with TraCS.}
        TraCS encodes natural-language descriptions of compliant road behavior via an agentic coding pipeline into probabilistic first-order logic programs.
        Based on TraCS' inferred level of compliance and contextual alignment, the backbone's set of predictions is reweighted to respect the environment's agreed-upon motion constraints.
    }
    \label{fig:tracs_overview}
\end{figure}

\section{Trajectory Compliance-Shaping for Motion Predictions}

To create TraCS, we begin by introducing a probabilistic, first-order vocabulary for spatial relations over agent states and state transitions. 
This allows us to leverage LLMs to translate natural-language descriptions of traffic regulations into interpretable probabilistic compliance programs. 
We then describe how inference in these programs allows TraCS to score candidate trajectories and introduce a neural calibration mechanism that governs when and how TraCS's beliefs reshape the backbone's original trajectory weights. 
The TraCS pipeline is illustrated in~\Cref{fig:tracs_overview}.

\subsection{Transitional Statistical Spatial Relations}
\label{sec:starmap}

Ingesting traffic rules in natural language to steer black-box motion predictors without retraining requires an interface sufficiently expressive to cover common traffic rules and exceptions.
Inspired by works on modeling uncertain environments in shared traffic spaces~\citep{DBLP:journals/firai/KartmannA23,flade2023star,DBLP:conf/iros/DoulaYMG24nesymof}, we employ a hybrid probabilistic first-order language to express traffic rules. 
More specifically, we follow and extend on the concept of Statistical Relational Maps (StaR Maps), which formalize reasoning over uncertain spatial relations in urban mobility~\citep{flade2023star}.
A StaR Map spatial relation is a function $r(\mathbf x, F)$ that maps a state $\mathbf x \in \mathbb R^D$ and a set of uncertain feature geometries $F$ to qualitative (e.g., "$\mathbf x$ is \textit{within} $F$") or quantitative (e.g., "$\mathbf x$ is $10m$ away from $F$") values.
By taking the $n$-th sample of each geometry in $F$, we obtain a set of $N$ randomly transformed samples $\mathcal F = \{ F^{(n)} \}_{n\in\{1,...,N\}}$. 
Interpreting the output of a spatial relation $r$ as outcomes of a probability distribution then allows for computing the statistical moments of said distribution using $\mathcal{F}$. 

For relation $r$ and a location $\mathbf x$, the parameters $\widehat{\bm{\theta}}_{r, \mathbf x}(\mathcal F) = (\widehat{\mu_r}, \widehat{\sigma^2_r})$ of $r$'s distribution can hereby be estimated.
In ground mobility, reasoning based only on locations is often ambiguous, e.g., when distinguishing between a vehicle speeding and one driving carefully. 
We therefore extend this framework towards Transitional StaR Maps by considering a state's first derivative $\dot{\mathbf x}\in \R^D$ and a time step size $\delta t \in \R^+$.
On this domain, we define transitional relations $r_{\delta t}(\mathbf x, \dot{\mathbf x}, F)$ such that
\begin{equation}
    s_{r_{\delta t}, \mathcal F}: \mathbb R^D \times \mathbb R^D\to \mathbb R^e, \ (\mathbf x, \dot{\mathbf x}) \mapsto \widehat{\bm\theta}_{r_{\delta t}, \mathbf x, \dot{\mathbf x}}(\mathcal F).
\end{equation}
calculates parameters using both $\mathbf x$ and $\dot{\mathbf x}$.
Reasoning over state transitions with Transitional StaR Maps thus allows for the definition of logical predicates that, e.g., depend on agent speed or restrict certain directed movements.
We define the considered relations employed for TraCS in Appendix~\ref{app:spatial}.

\subsection{Agentic Traffic Regulation Modeling}

Given a natural-language description of the applicable traffic regulations, whether user-specified or sourced directly from an official legal document, TraCS employs an agentic LLM pipeline to produce a formal, executable compliance model $\mathcal{R}$ including a compliance clause $C$.
Concretely, the regulations are encoded as a probabilistic logic program in Resin~\citep{DBLP:journals/corr/kohautReactive}, expressed using a vocabulary of probabilistic spatial relations over agent trajectories as defined above. 
The LLM agents are thereby enabled to encode complex point-wise and transitional constraints over agent trajectories.

Our pipeline is an iterative refinement loop around three roles.
First, a \textit{coder} agent is prompted with the natural-language regulation description and the available spatial relations to generate a candidate Resin program.
Second, a \textit{judge} agent independently evaluates whether the program faithfully represents the intent of the description, identifying logical gaps or misrepresentations.
Third, the Resin compiler acts as a \textit{verifier tool} to ensure the program is well-formed.

If either the judge deems the program non-representative or the compiler rejects it, the resulting feedback, comprising the judge's critique and any compiler error messages, is returned to the coder agent, which is instructed to revise its output accordingly. 
This cycle repeats until both the judge and the compiler are satisfied, at which point the program is accepted as the traffic compliance model for the current scenario. 
The iterative, multi-role structure of this pipeline draws inspiration from recent work on agentic code generation~\citep{DBLP:conf/kr/IshayY023generateASP}, and ensures that the resulting compliance model is both semantically grounded in the original regulations and formally executable.

\subsection{Neuro-Symbolic Trajectory Scoring}
\label{sec:fusion}

TraCS determines for each candidate trajectory $\bm{\tau}_k \!=\! (\mathbf{x}_1, \dots, \mathbf{x}_T)$ the probabilities $P(C\!\mid\! \mathbf{x}, \dot{\mathbf x}, \delta t, \mathcal{R})$, i.e., the probability that across $\bm{\tau}_k$ each $\mathbf{x}$ and its transition into the next state satisfy the compliance clause $C$ defined in the rules $\mathcal{R}$.
To evaluate this quantity, one must assign to each logical atom $a \in \mathcal{A}$ appearing in $\mathcal{R}$ the probability $P(a = j(a) \mid \mathbf{x}, \dot{\mathbf x}, \delta t, \mathcal{R})$ of atom $a$ taking on the value assigned by grounding model $j$. 
The atom probabilities, in turn, are provided by estimating a Transitional StaR Map for the environment.
The resulting point-wise compliance score is then obtained via weighted model counting over the set of all models $\mathcal{J}$ of the program:
\begin{align}
    P(C \mid \mathbf{x}, \dot{\mathbf x}, \delta t, \mathcal{R}) = \sum\nolimits_{j \in \mathcal{J}} \prod\nolimits_{a \in \mathcal{A}} P(a = j(a) \mid \mathbf{x}, \dot{\mathbf x}, \delta t, \mathcal{R}).
    \label{eq:wmc}
\end{align}
Intuitively, each model $j \in \mathcal{J}$ corresponds to a complete truth assignment over the atoms of $\mathcal{R}$, and its weight is the product of the individual atom probabilities under that assignment. 
Summing over all satisfying models then yields the total probability of compliance, marginalizing over all consistent groundings of the program. 
We implement this step using Resin and its Reactive Circuits~\citep{DBLP:journals/corr/kohautReactive} as runtime-efficient inference engines on top of Answer Set Programming~\citep{DBLP:journals/tplp/GebserKKS19}.
To obtain TraCS probabilities for a candidate trajectory, we compute the geometric mean of the pointwise probabilities and normalize after:
\begin{align}
    p_k^{\text{symbolic}} \propto \exp\!\left( \frac{1}{N} \sum_{n=1}^{N} \log P(C \mid \mathbf{x}, \dot{\mathbf{x}}, \delta t, \mathcal{R}) \right).
    \label{eq:traj_score}
\end{align}
This formulation treats compliance as a property that must hold consistently across the trajectory, such that a single severely non-compliant state exerts a disproportionate downward pull on the overall score, reflecting the intolerance of traffic regulations for isolated violations.
Given the original distribution of probabilities $p_k^\text{neural}$ of a motion forecasting backbone and the probabilities $p_k^{\text{symbolic}}$ of TraCS, we apply a log-linear opinion pool where the symbolic log-probability is centered against the log-uniform baseline $\log(\nicefrac{1}{6})$, such that the symbolic term vanishes when $p_k^{\text{symbolic}}$ is uninformative and grows with its deviation from uniformity:
\begin{align}
    p_k^{\text{fused}} \propto \exp\left(\log{p_k^{\text{neural}}} + w \cdot \left(\log{p_k^{\text{symbolic}}} - \log \frac{1}{6}\right)\right) .
    \label{eq:log_pool}
\end{align}
Consequently, if all candidates of an intentionally non-compliant agent violate $\mathcal{R}$, they receive comparably low compliance, $p_k^{\text{symbolic}}$ approaches uniformity, and the fused distribution collapses back to the backbone.

\subsection{Model Alignment and Trust Gating}

A key challenge in applying symbolic compliance scoring to motion prediction is that the logic program $\mathcal{R}$, however carefully constructed, is an imperfect model of real-world traffic behavior. 
In certain environmental contexts, e.g., characterized by factors such as road geometry, agent velocities, or local traffic patterns, the compliance scores produced by \Cref{eq:traj_score} may systematically differ from observed ground-truth trajectories, reflecting gaps or oversimplifications in the encoded regulations.
We therefore fit a backbone-specific trust network that emits a trust score $\hat{w}\in[0,1]$:
\begin{align}    
\hat{w} \;=\; \sigma\big(\,\textsc{MLP}_{\text{trust}}\big(
   [\,\mathbf{x}^{\text{feat}}\,;\;\textsc{MLP}_{\text{emb}}(\mathbf{x}^{\text{emb}})\,]
\big)\,\big).
\end{align}
Here, we combine a set of handcrafted interpretable features $\mathbf{x}^{\text{feat}}$ together with the backbone's respective environment embeddings $\mathbf{x}^{\text{emb}}$ to guide the network's decision.
For each scenario in the training fold, we form a per-scenario regression target $w^{\star}\!\in\![0,1]$ by searching the scalar fusion weight that minimizes the expected Average-Displacement Error (ADE) under the fused distribution:
\begin{align}
  w^{\star}
   \;=\; \argmin_{w \in [0,1]}\;
         \sum_{k=1}^{6} p^{\text{fused}}_k(w)\,\cdot\,\text{ADE}(\bm{\tau}_k, \bm{\tau}^\star).
\end{align}
The trust network is then fitted with Binary Cross-Entropy on the soft target
$w^{\star}$:
\begin{align}
\mathcal{L}(\hat{w},w^{\star})
   \;=\;
   -\,w^{\star}\log \hat{w}\;-\;(1-w^{\star})\log(1-\hat{w}).
\end{align}
Since $w^\star$ minimizes expected displacement error, the gate learns to abstain wherever compliant behavior is unlikely.

Although $\hat{w}$ captures a degree of trust in the fit of TraCS in a given scenario, we introduce an additional trust threshold $\theta$ in order to avoid degradations from reshaping backbone predictions at low certainty. 
Hence, we apply a binary gate at a fixed threshold $\theta \in [0,1]$: $w_{\text{gate}}(\hat{w};\theta) = 1$ if $\hat{w} \ge \theta$, and $0$ if $\hat{w} < \theta$.
Thereafter, we pass $w_{\text{gate}}$ into the log-linear opinion pool of \Cref{eq:log_pool}. 
To select the threshold $\theta$ \emph{per backbone} as part of the training, we sweep $[0, 1]$ and choose the value that minimizes validation bMinADE$_1$~\citep{Argoverse2} as a probabilistic calibration target.

\section{Experiments}
\label{sec:experiments}
We evaluate TraCS on the Argoverse~2 (AV2) motion-forecasting benchmark~\citep{Argoverse2}, which asks for $K{=}6$ future trajectory candidates of each focal agent over a six-second horizon. 
We instantiate TraCS atop three publicly available backbones spanning the recent state of the art: \textsc{DeMo}~\citep{DBLP:conf/nips/ZhangSZ24demo}, \textsc{Polaris}~\citep{zhang2026polaris}, and \textsc{QCNet}~\citep{DBLP:conf/cvpr/Zhou0LH23qcnet}. 
For every scenario, TraCS scores each of the backbone's six candidates against the same compiled logic program (shown in Appendix~\ref{app:agentic}), and a gated trust network emits a per-scenario trust decision $w \in \{0,1\}$ that activates fusion only when TraCS is likely to be informative. 
\begin{wrapfigure}{r}{0.5\textwidth}
    \centering
    \includegraphics[width=0.74\linewidth]{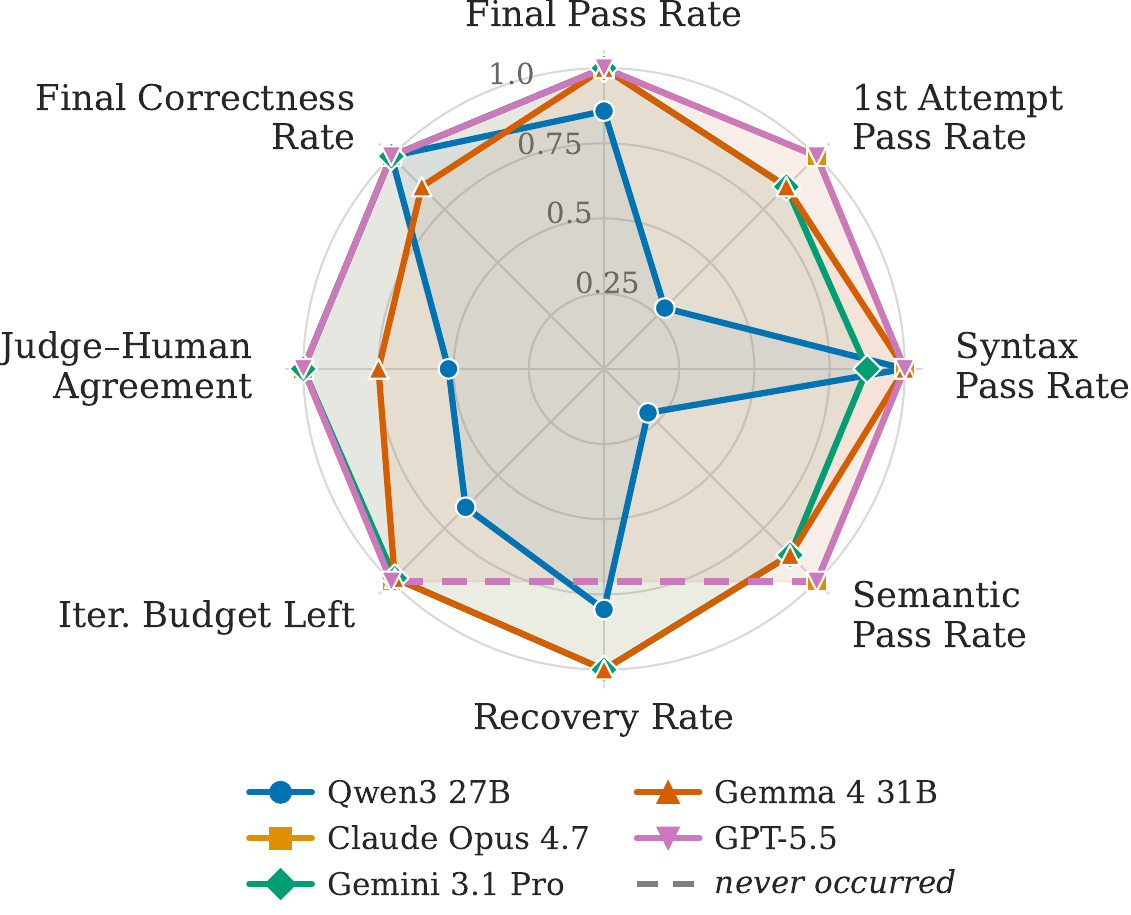}
    \caption{
        \textbf{Agentic traffic modeling.}
        Across open-weight and foundation models, not all are able to produce the correct syntax and semantics.
    }
    \label{fig:agent_capabilities}
    \vspace{-2\baselineskip}
\end{wrapfigure}
The gate threshold $\theta$ is selected per backbone on a held-out validation split.
The detailed machine learning setup is given in Appendix~\Cref{app:ml_setup}. 
All results report the mean over 10 random seeds and stem from the AV2 single-agent motion forecasting test split ($n{=}24{,}984$ focal agents).
Compiling the logic program takes on average $21.04\,$s and $5998$ tokens; inference runs at $1.86\!\pm\!0.29\,$Hz and fusion at $6060\,$Hz.

\subsection{Agentic Codification for TraCS}
We test the codification quality across open-weights and proprietary foundation models.
Results are shown in \Cref{fig:agent_capabilities}.
For the following TraCS experiments, we employ a conjunction of all encoded programs (i.e., over individual compliance constraints).
The system prompts, judge prompt, and generated programs (focused on underrepresented classes from AV2) are shown in Appendix~\ref{app:agentic}.

\begin{table}[t]
\centering
\caption{Effect of TraCS-gated fusion on Argoverse 2 Brier-style metrics across backbones and modes of mobility. 
Relative change is shown next to each +TraCS value, coloured by sign: \textcolor{impcol}{green} = improvement, \textcolor{degcol}{red} = degradation, \textcolor{neucol}{grey} = negligible ($|\Delta| < 0.05\%$). 
Lower is better for all metrics  (\textdownarrow).}
\label{tab:brier_table}
\footnotesize
\setlength{\tabcolsep}{5pt}
\renewcommand{\arraystretch}{1.1}
\begin{tabular}{l l r r r r r r}
\toprule
 & & \multicolumn{2}{c}{bMinADE$_6$  (\textdownarrow)} & \multicolumn{2}{c}{bMinADE$_1$  (\textdownarrow)} & \multicolumn{2}{c}{bMinFDE$_1$ (\textdownarrow)} \\
\cmidrule(lr){3-4}\cmidrule(lr){5-6}\cmidrule(lr){7-8}
Backbone & Mode & Base & +TraCS & Base & +TraCS & Base & +TraCS \\
\midrule
\multirow{6}{*}{DeMo}
 & Overall      & 1.3185 & 1.3172\dgood{0.1} & 2.1900 & 2.1842\dgood{0.3} & 4.5940 & 4.5887\dgood{0.1} \\
 & Bus          & 1.5107 & 1.5073\dgood{0.2} & 2.5072 & 2.4661\dgood{1.6} & 4.8436 & 4.7718\dgood{1.5} \\
 & Cyclist      & 1.3356 & 1.3305\dgood{0.4} & 1.9574 & 1.9493\dgood{0.4} & 3.7022 & 3.7014\dflat{0.0} \\
 & Motorcyclist & 1.5473 & 1.5450\dgood{0.1} & 2.6472 & 2.6258\dgood{0.8} & 5.3926 & 5.3562\dgood{0.7} \\
 & Pedestrian   & 1.0041 & 1.0016\dgood{0.2} & 1.2375 & 1.2121\dgood{2.1} & 1.9985 & 1.9690\dgood{1.5} \\
 & Vehicle      & 1.3318 & 1.3307\dgood{0.1} & 2.2468 & 2.2440\dgood{0.1} & 4.7807 & 4.7800\dflat{0.0} \\
\midrule
\multirow{6}{*}{Polaris}
 & Overall      & 1.2713 & 1.2696\dgood{0.1} & 2.1131 & 2.1075\dgood{0.3} & 4.4036 & 4.4043\dflat{0.0} \\
 & Bus          & 1.4685 & 1.4680\dflat{0.0} & 2.4212 & 2.4074\dgood{0.6} & 4.6466 & 4.6806\dbad{0.7}  \\
 & Cyclist      & 1.2977 & 1.2957\dgood{0.2} & 1.8204 & 1.8187\dgood{0.1} & 3.3874 & 3.3959\dbad{0.3}  \\
 & Motorcyclist & 1.4241 & 1.4172\dgood{0.5} & 2.3567 & 2.3432\dgood{0.6} & 4.7648 & 4.7657\dflat{0.0} \\
 & Pedestrian   & 0.9565 & 0.9534\dgood{0.3} & 1.1383 & 1.1218\dgood{1.4} & 1.8903 & 1.8736\dgood{0.9} \\
 & Vehicle      & 1.2846 & 1.2830\dgood{0.1} & 2.1740 & 2.1695\dgood{0.2} & 4.5890 & 4.5895\dflat{0.0} \\
\midrule
\multirow{6}{*}{QCNet}
 & Overall      & 1.2477 & 1.2476\dflat{0.0} & 2.0693 & 2.0643\dgood{0.2} & 4.6856 & 4.6765\dgood{0.2} \\
 & Bus          & 1.4119 & 1.4147\dbad{0.2}  & 2.3010 & 2.2628\dgood{1.7} & 4.7257 & 4.6768\dgood{1.0} \\
 & Cyclist      & 1.2634 & 1.2657\dbad{0.2}  & 1.7129 & 1.7020\dgood{0.6} & 3.4995 & 3.4858\dgood{0.4} \\
 & Motorcyclist & 1.3736 & 1.3747\dbad{0.1}  & 2.2972 & 2.2786\dgood{0.8} & 5.2616 & 5.2346\dgood{0.5} \\
 & Pedestrian   & 0.9522 & 0.9522\dflat{0.0} & 1.0634 & 1.0586\dgood{0.5} & 1.8676 & 1.8621\dgood{0.3} \\
 & Vehicle      & 1.2612 & 1.2610\dflat{0.0} & 2.1367 & 2.1332\dgood{0.2} & 4.9029 & 4.8953\dgood{0.2} \\
\bottomrule
\end{tabular}
\end{table}

\subsection{Motion Forecasting on TraCS}

\autoref{tab:brier_table} reports the Argoverse~2 Brier-style metrics with and without TraCS, broken down overall and by focal-agent class. 
Across all three backbones, gated TraCS fusion produces a monotone improvement on the headline bMinADE$_1$ metric; every $(\text{backbone}\!\times\!\text{modality})$ cell improves. 
Gated fusion improves bMinADE$_1$ by $-0.0039$, $-0.0061$, and $-0.0047$ for \textsc{DeMo}, \textsc{Polaris}, and \textsc{QCNet}; a two-sided $t$-test on the per-seed means rejects $H_0\!:\!\Delta{=}0$ for all three (Holm-corrected $p \leq 2.0\!\times\!10^{-4}$), and hierarchical bootstrap 95\% CIs over scenarios and seeds exclude zero.
Aggregating per class to give each population stratum equal weight yields balanced improvements of $-0.0145$, $-0.0100$, and $-0.0173$ ($p\!\leq\!2.7\!\times\!10^{-5}$), improving in every run.

bMinFDE$_1$ improves consistently on \textsc{DeMo} and \textsc{QCNet} across all classes; 
on \textsc{Polaris} it improves substantially on the under-represented pedestrian class ($-0.9\%$) but shows small per-class regressions of at most $+0.7\%$ on bus and cyclist, which we attribute to that backbone's calibration already concentrating mass on favourable endpoints. 

The $K{=}1$ deterministic point metrics (minADE$_1$, minFDE$_1$, MR$_1$),
omitted from \autoref{tab:brier_table} for brevity, remain within $\pm 0.3\%$ of the backbone at the population level. 
They characterize a calibration-to-geometry trade-off: by sharpening the predictive distribution onto TraCS's preferred candidate, fusion occasionally selects a trajectory that is marginally less geometrically accurate but more confidently endorsed. 
The trade is favorable in expectation under the AV2 Brier criterion and allows treating some of the most severe mispredictions, which we will investigate in more detail next.

\subsection{TraCS-based Tail-Risk Mitigation}
\label{sec:exp:tail}

\begin{figure}[t]
  \centering
  \includegraphics[width=\textwidth]{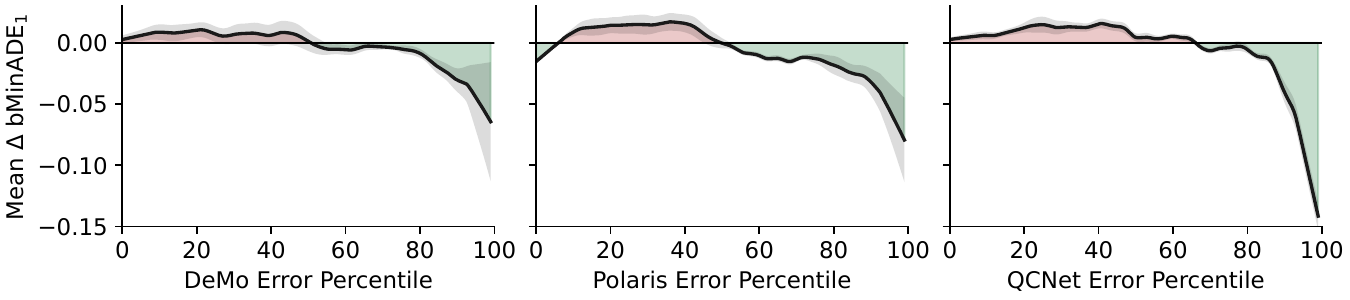}
  \caption{
        \textbf{TraCS as a tail-risk treatment:} 
        Smoothed (LOWESS) conditional mean of $\Delta$ bMinADE$_1$ (\textdownarrow{} is better) per scenario as a function of backbone-error percentile, for each backbone.
        Filled regions encode the sign of the smoothed mean: green where fusion improves, red where it regresses. 
        The shape is consistent across backbones; a flat plateau through the lower 60--70\% of scenarios and a dip in the upper decile, where the backbone is least reliable.
    }
    \label{fig:percentile_difference}
\end{figure}

\begin{figure}
    \centering
    \includegraphics[width=\linewidth]{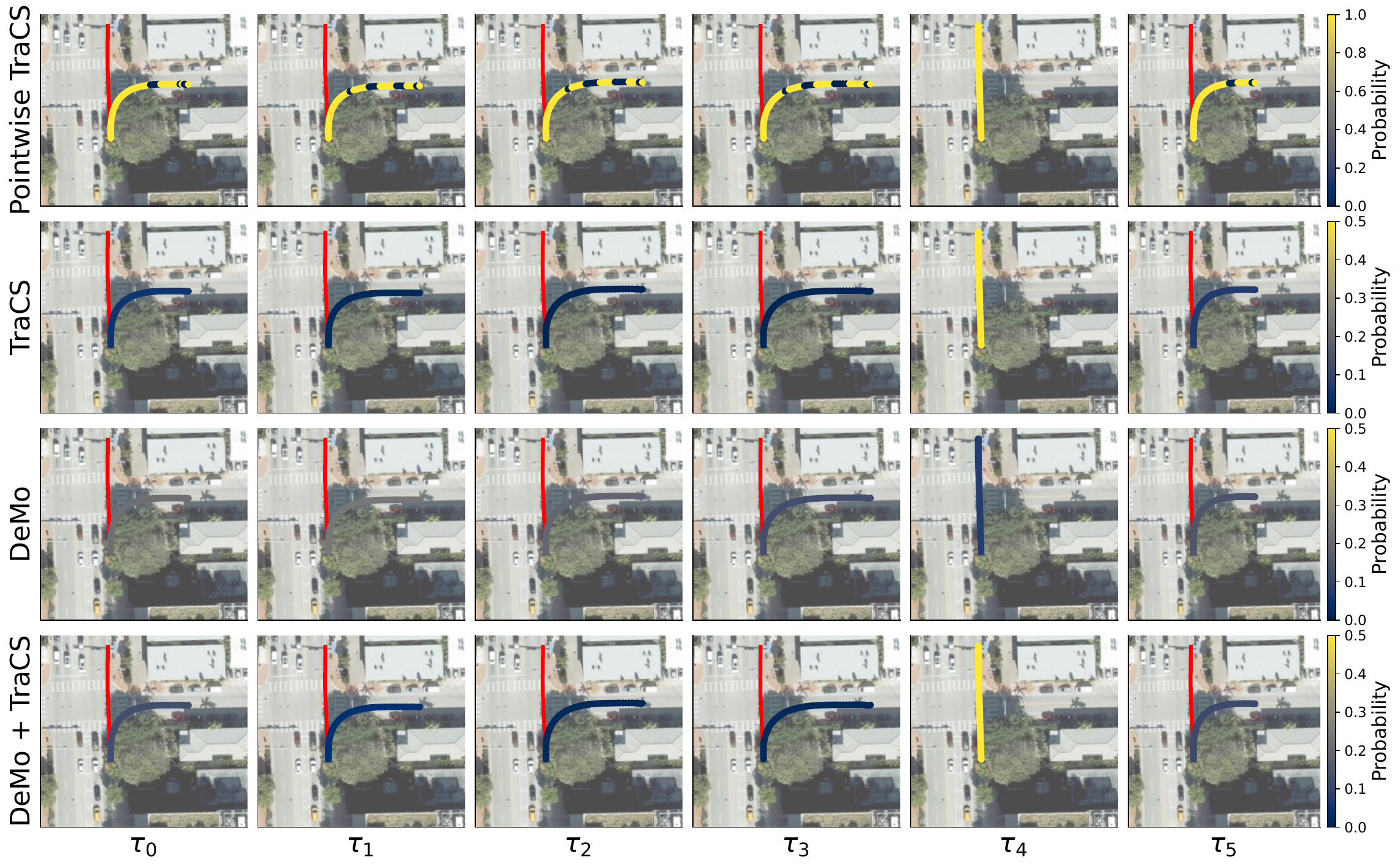}
    \caption{
        \textbf{TraCS shaping a backbone prediction:}
        Here, a bus proceeds straight through a four-way intersection (\textcolor{red}{red} ground-truth trajectory).
        While the backbone (here, DeMo) has no clear preference on any of its proposal trajectories, TraCS correctly recognizes unlikely or incompliant behavior and shifts the probability mass towards the correct proposal, promoting $\tau_4$ from being the \textcolor{blue}{lowest} to being the \textcolor{Goldenrod}{highest} rated one. 
        Note that the color scale differs between pointwise and per-trajectory scores.
    }
    \label{fig:case_study}
\end{figure}

We show the conditional mean of $\Delta$ bMinADE$_1$ as a function of the scenario's \emph{backbone-error percentile} in \autoref{fig:percentile_difference}. 
Across all three backbones, the mean change is essentially zero on the lower $\sim$60-70\% of scenarios, where the backbone is already strong, and dips sharply into the green region within the top quintile, with the worst-decile mean change reaching several times the population mean.
The worst-decile gain ($-0.044$, $-0.054$, $-0.088$) is $8.9$--$18.7\times$ the population mean and significantly exceeds that of the easier half (interaction $p\!\leq\!1.2\!\times\!10^{-3}$).
Gated TraCS fusion therefore functions as a \emph{tail-risk treatment} for the predictor, leaving typical predictions untouched and selectively repairing the worst-case scenarios that would dominate downstream planner risk.
\autoref{fig:case_study} illustrates the qualitative effect of TraCS gating on a single \textsc{DeMo} scenario drawn from the high-error tail.

This view also reconciles the apparent tension between the small population-mean improvement and the dramatic per-scenario rescue in \autoref{fig:case_study}: 
The test population is dominated by scenarios in which the gate correctly abstains, so a single-number average understates the policy's value.


\subsection{Impact per Mode of Mobility}
\label{sec:exp:distribution}

\begin{figure}[t]
  \centering
  \includegraphics[width=\textwidth]{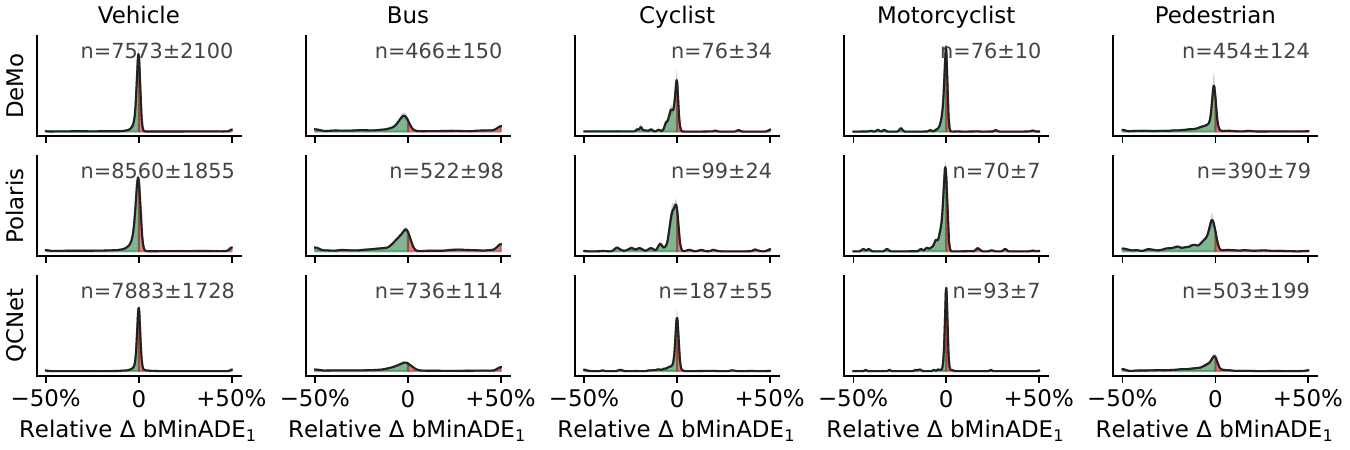}
  \caption{\textbf{TraCS improves prediction calibration across modes of mobility:}  
    We show kernel density estimates of the difference between bMinADE$_1$ after applying TraCS.
    Mass left of zero is improvement (green), mass right of zero is regression (red); $n$ denotes the sample size for which TraCS was applied ($w_\text{gate}=1$).
    Focal agents that are more clearly constrained by the employed logic exhibit more left-skewed distributions, while others concentrate tightly around zero, indicating that TraCS's gated fusion correctly leaves predictions intact where TraCS provides no information gain.
  }
  \label{fig:modality_wise_impact}
\end{figure}

\autoref{fig:modality_wise_impact} shows the per-scenario \emph{relative} change in bMinADE$_1$ \emph{conditional on the gate firing} ($w_\text{gate}{=}1$), distributed by the focal agent's mode of mobility (vehicle, bus, cyclist, motorcyclist, pedestrian) and stratified by backbone. 
Conditioning on $w_\text{gate}{=}1$ isolates the behavior of fusion when it is actually applied in contrast to the population-level effect shown previously in~\autoref{fig:percentile_difference}.
Two observations are worth highlighting. 

\textbf{(i)} The infrequently-observed classes, i.e., buses, cyclists, and pedestrians, show the most left-skewed distributions, with medians substantially below zero and a long tail of large improvements. 
These are exactly the classes for which a backbone trained on a vehicle-heavy distribution is least reliable and for which a logically-grounded fallback adds the most value.
\textbf{(ii)} For vehicles, which dominate the dataset, the distribution is narrowly peaked around zero with a near-balanced sign profile.
Hence, TraCS neither systematically helps nor hurts the class that the backbones already handle well. 

\subsection{Ablation Studies}
\autoref{tab:ablations} isolates the effect of different gating mechanisms.
Always fusing degrades every backbone, so gating recovers $0.035$, $0.090$, and $0.014$ bMinADE$_1$ (paired $p\!<\!10^{-11}$).
Shuffling the TraCS scores across scenarios while holding the gate decisions fixed renders fusion harmful again, attributing the gain to the encoded logic rather than to reweighting per se.
While a gate without backbone-specific latents using only $\mathbf{x}^{\text{feat}}$ may perform on par with the full gate, transferring a latent-free gate zero-shot to another backbone is not reliably beneficial (\autoref{tab:gate_transfer}), i.e., the gate must still be fitted per backbone.
   
\begin{table}[t]
\centering
\small
\setlength{\tabcolsep}{4pt}
\renewcommand{\arraystretch}{1.15}
\begin{minipage}[t]{0.5\linewidth}
  \centering
  \caption{Mean $\Delta\text{bMinADE}_1$ comparisons against the unmodified backbone (lower is better).}
  \label{tab:ablations}
  \begin{tabular}{lrrr}
    \toprule
    Gate policy & DeMo (\textdownarrow) & Polaris (\textdownarrow) & QCNet (\textdownarrow) \\
    \midrule
    Always fuse            & $+0.0312$ & $+0.0842$ & $+0.0088$ \\
    Shuffled TraCS         & $+0.0149$ & $+0.0252$ & $+0.0035$ \\
    Latent-free gate       & $-0.0038$ & $-0.0051$ & $-0.0056$ \\
    \textbf{TraCS (ours)}  & $-0.0039$ & $-0.0061$ & $-0.0047$ \\
    \midrule
    Oracle gate            & $-0.0808$ & $-0.1199$ & $-0.0425$ \\
    \bottomrule
  \end{tabular}
\end{minipage}\hfill
\begin{minipage}[t]{0.45\linewidth}
  \centering
  \caption{Mean $\Delta\text{bMinADE}_1 \times10^{-3}$ (zero-shot latent-free gate transfer) against the unmodified backbone (lower is better).}
  \label{tab:gate_transfer}
  \begin{tabular}{lrrr}
    \toprule
    & \multicolumn{3}{c}{Eval (\textdownarrow)} \\
    \cmidrule(lr){2-4}
    Train & DeMo & Polaris & QCNet \\
    \midrule
    DeMo    & \cellcolor{black!8}$-3.8$\std{1.7} & $+0.2$\std{4.3}   & $-1.4$\std{1.3} \\
    Polaris & $-3.2$\std{1.4} & \cellcolor{black!8}$-5.1$\std{1.8} & $-0.7$\std{0.7} \\
    QCNet   & $+7.9$\std{4.4} & $+29.2$\std{16.4} & \cellcolor{black!8}$-5.6$\std{2.0} \\
    \bottomrule
  \end{tabular}
\end{minipage}
\end{table}

\section{Conclusion}

\paragraph{Summary.}
We have presented TraCS, a neuro-symbolic framework for motion prediction in multi-modal ground mobility environments, spanning agents such as cars, bicycles, and buses. 
Our method is designed as a plug-and-play compliance density module compatible with existing motion prediction frameworks,  augmenting their outputs without requiring architectural changes. 
At its core, an LLM generates a declarative first-order logic program encoding the compliant behavior of each agent class within a regulated traffic environment. 
To ground this symbolic reasoning in real-world uncertainty, we apply statistical modeling over the environment to produce hybrid probabilistic ground atoms, serving as the symbolic vocabulary for the LLM's logical description. 
By learning a neural trust alignment that contextually weights TraCS' influence, we control the posterior's steering in situations where the symbolic model may not faithfully model the focal agent's actual decisions. 
We show that our setup consistently improves state-of-the-art motion forecasting for multi-modal ground mobility, demonstrating that embedding structured compliance reasoning atop neural representations yields critical gains at the tail end.

Beyond its immediate application to motion prediction, TraCS illustrates how natural-language regulations can be transformed into interpretable probabilistic reasoning processes that influence autonomous behavior. 
While demonstrated in traffic environments, this general paradigm may also support social simulations in which humans, robots, and AI agents interact under shared normative structures. 
Neuro-symbolic representations of regulations and conventions could thereby provide an important foundation for studying socially situated autonomous systems beyond dyadic interactions.

\paragraph{Limitations and Future Work.}
Our proposed framework treats the probabilistic logic program as an inference-time artifact, regenerated via LLM prompting rather than optimized during end-to-end training. 
As a consequence, the conditional probabilities governing individual rule activations are not learned from data, leaving information on the table that a differentiable or traditional EM-style update scheme may leverage. 
A natural extension is to close this loop by using the probabilistic logic structure as a differentiable inductive bias through which neural sensor representations could be learned end-to-end, transitioning the method from a post-hoc overlay to an integral component of tuning an agent's prediction pipeline. 
Additionally, the present model treats agents as conditionally independent given the scene context, whihout explicitly representing joint maneuvers or negotiations between agents. 
Modeling such inter-agent dependencies is particularly consequential at longer prediction horizons, where the compounding effects of agent interactions induce additional uncertainty.
Extending the framework toward such interaction-dependent reasoning may also provide a bridge from motion forecasting to more general settings. 
In hybrid environments involving both humans and robots, behavior emerges through repeated interactions under shared regulations, social conventions, and collective agreements. 
Representing these normative structures in executable probabilistic logic could enable the systematic study of how local behavioral constraints shape emergent behavior beyond dyadic settings. 
Traffic environments provide a structured starting point for developing these mechanisms, while future extensions could examine their applicability to community-level hybrid human-AI societies.

\section*{Acknowledgements}

Simon Kohaut gratefully acknowledges the financial support from Honda Research Institute Europe~(HRI-EU).
This work received funding from the German Research Foundation~(DFG) through the Priority Program SPP 2422 within the subproject \enquote{KIWi-SP}~(KE 1686/10-1).
The work has also benefited from the ICT-48 Network of AI Research Excellence Center \enquote{TAILOR} (EU Horizon 2020, GA No 952215) and the Excellence Cluster EXC-3057, funded by the DFG under Germany's Excellence Strategy.
The TU Eindhoven authors received support from their Department of Mathematics and Computer Science and the Eindhoven Artificial Intelligence Systems Institute.
Argoverse 2 data \copyright~Argo AI LLC, licensed under the \href{https://creativecommons.org/licenses/by-nc-sa/4.0/legalcode}{CC BY-NC-SA 4.0} and available from \href{https://www.argoverse.org/av2}{argoverse.org/av2}.

\bibliography{main}

\clearpage
\appendix
\section{Declarative Traffic Modeling}
As introduced in section \ref{sec:starmap}, we use a domain-specific language to reason about constraints that apply to traffic participants in ground mobility scenarios. 
In the following, detailed mathematical definitions of the introduced concepts are provided, along with concrete examples of the relations used in the experiments.

\subsection{Transitional and Hybrid Probabilistic Spatial Relations}
\label{app:spatial}

A transitional spatial relation has been defined as a function $r_{\delta t} : \R^D \times \R^D \times \mathbb F \to \B \text{ or } \R$ over locations $\mathbf x \in \R^D$, derivatives $\dot{\mathbf x}\in\R^D$ and a set of feature geometries $F\in \mathbb F$. For binary relations, the codomain is $\B$, and for continuous ones it is $\R$.

In our implementation, a set of feature geometries $F = (M, f)$ consists of a graph-based map $M$ and a feature type $f \in \mathcal T$. A map $M = (\mathcal{V}, \mathcal{E}, \rho)$ is triple of position vertices $\mathcal{V}$, edges $\mathcal{E}$ between them, and a tagging function $\rho : \mathcal{V} \to \mathcal{P}(\mathcal{T})$.
The function $\rho(\bm{v}) \subseteq \mathcal{T}$ annotates each vertex $\bm{v} \in \mathcal{V}$ with a set of semantic tags, such as a feature type $f \in \mathcal{T}$. If a path exists between two vertices in $\mathcal{V}$ across edges in $\mathcal{E}$, they are considered parts of the same geometry.

The introduction of transitional semantics enables extending the standard vocabulary of StaR Maps, consisting of point-based relations such as \textit{over} and \textit{distance}, by defining the transitional spatial relations \textit{intersects}, \textit{crosses}, \textit{enters}, \textit{exits}, \textit{approaches}, \textit{follows}, and \textit{opposes}.

Concretely, we define
\begin{align}
    \text{intersects}_{\delta t}(\mathbf{x}, \dot{\mathbf{x}}, (M, f)) := \begin{cases}
        1 & \text{if }
            \{ \mathbf{x} + \lambda\dot{\mathbf{x}} \mid \lambda \in [0, \delta t] \}
            \cap
            \mathcal{A}(M_f) \neq \emptyset \\
        0 & \text{else}
    \end{cases} ,
\end{align}
where $M_f$ is the map filtered for type $f$, i.e., $M_f := (\mathcal{V}_f, \mathcal{E} \cap \mathcal{V}_f^2, \rho)$, with vertices $\mathcal{V}_f := \{\bm v \mid f \in \rho(\bm v)\}$. Furthermore, the area $\mathcal{A}(M)$ is defined as the set of all points covered by the geometries in the map $M$.
For \textit{crosses}, we first define the set of inner points of the movement prediction $\xi = \{ \mathbf{x} + \lambda\dot{\mathbf{x}} \mid \lambda \in (0, \delta t) \}$ and the set of all maps of connected components $\operatorname{Comp}(\mathcal{M}) = \left\{ (\mathcal{V}', \mathcal{E}', \rho ) \mid \mathcal{V}' \subseteq \mathcal{V} \land \mathcal{E}' = \mathcal{E} \cap \mathcal{V}'^2 \land \mathcal{V}' \text{ connected component in } (\mathcal{V}, \mathcal{E}) \right\}$. Then,
\begin{align}
    \text{crosses}_{\delta t}(\mathbf{x}, \dot{\mathbf{x}}, (M, f)) &:= \begin{cases}
        1 & \text{if } \,\,
        \begin{aligned}[t]
            & \exists m \in \operatorname{Comp}(M_g): \\
            & \xi \cap \mathcal{A}(m) \neq \emptyset \;\land\; \xi \neq \xi \cap \mathcal{A}(m) \neq \mathcal{A}(m) \;\land \\
            & \!\operatorname{dim}(\xi \cap \mathcal{A}(m)) = \operatorname{dim}(\mathcal{A}(m))-1
        \end{aligned} \\
        0 & \text{else }
    \end{cases}.
\end{align}
With the time step $\delta t \in \R_+$, we can extrapolate $\mathbf{x}$ to $\mathbf{x}' := \mathbf{x} + \delta t \dot{\mathbf{x}}$ under a simple forward model.
Using this predicted $\mathbf{x}'$, we can define
\begin{align}
    \text{enters}_{\delta t}(\mathbf{x}, \dot{\mathbf{x}}, (M, f)) &:= \begin{cases}
        1 & \text{if }
            \mathbf{x} \not\in \mathcal{A}(M_f) \land \mathbf{x}' \in \mathcal{A}(M_f) \\
        0 & \text{else}
    \end{cases}, \\
    \text{exits}_{\delta t}(\mathbf{x}, \dot{\mathbf{x}}, (M, f)) &:= \begin{cases}
        1 & \text{if }
            \mathbf{x} \in \mathcal{A}(M_f) \land \mathbf{x}' \not\in \mathcal{A}(M_f) \\
        0 & \text{else}
    \end{cases} , \text{ and} \\
    \text{approaches}_{\delta t}(\mathbf{x}, \dot{\mathbf{x}}, (M, f)) &:= \begin{cases}
        1 & \text{if }
            \min_{\bm{v} \in \mathcal{V}_f } \norm{ \mathbf{x} - \bm{v} }_2 > \min_{\bm{v} \in \mathcal{V}_f } \norm{ \mathbf{x}' - \bm{v} }_2 \\
        0 & \text{else}
    \end{cases} .
\end{align}
For \textit{follows} and its opposite, \textit{opposes}, let $\operatorname{next}(\bm{v})$ denote the next point after $\bm{v}$ on a central line geometry through feature $f$, or $\bm{v}$ if it is the last one.
For line strings, this center line is just the geometry itself.
Intuitively, \textit{follows} takes the nearest line segment of the closest geometry in $M_f$ to $\mathbf{x}$, and calculates the alignment of that segment with $\dot{\mathbf{x}}$.
Formally, we compute
\begin{align}
    \text{follows}_{\delta t}(\mathbf{x}, \dot{\mathbf{x}}, (M, f)) &:=
    \min \left( 0,\ 
    \frac{ (\operatorname{next}(\bm{v}') - \bm{v}')^T \, \dot{\mathbf{x}}}{\norm{\operatorname{next}(\bm{v}') - \bm{v}'}_2 \cdot \norm{\dot{\mathbf{x}}}_2}
    \right)
    \quad\text{and} \\
    \text{opposes}_{\delta t}(\mathbf{x}, \dot{\mathbf{x}}, (M, f)) &:= \text{follows}_{\delta t}(\mathbf{x}, - \dot{\mathbf{x}},(M, f))
    ,
\end{align}
where $\bm{v}' := \argmin_{\bm{v} \in \mathcal{V}_f } \norm{ \mathbf{x} - \bm{v} }_2$ is the closest vertex $\bm v$ fro $\bm x$ in $M_f$.

\subsection{Agentic Coding Pipeline}
\label{app:agentic}

\begin{llmprompt}{System Prompt for the Coding Agent}
You are an expert logic programming coder specializing in traffic rules in the new Resin language.\\

YOUR TASK:\\
Translate natural language descriptions of traffic rules into strictly valid Resin code using world knowledge.
Resin programs encode first-order logic rules into the final landscape predicate.
Your task is to write the rules based on the natural language descriptions provided in the user input.
Comment lines start with a "\#". Use comments to clarify your code.\\

AVAILABLE ATOMS AND PREDICATES:\\
You can use the following atoms and predicates in your logic rules. They will be automatically populated with real-time data when the Resin program is executed. Feel free to derive new predicates as needed, but only based on the following atoms and existing rules.\\

Available map feature types are:\\
- Road: drivable\_area (union of all lanes).\\
- Road lanes: vehicle\_lane (open to all), bus\_lane, bike\_lane.\\
- Road lane center lines: vehicle\_lane\_centerline (open to all), bus\_lane\_centerline, bike\_lane\_centerline.\\
- Centerline of the lane of the focal agent: focal\_lane (>100 if unknown).\\
- Lane markings: solid\_line, dashed\_line.\\
- Special features: intersection, crossing.\\
- Moving traffic participants: vehicle, bus, cyclist, motorcyclist, pedestrian.

Available predicates, following Shapely semantics, for each feature type "X" are:\\
- over(X) (boolean)\\
- distance(X) (number, in meters)\\
- approaches(X) (boolean, the distance to X is decreasing)\\
- enters(X) (boolean, currently entering X)\\
- exits(X) (boolean, currently leaving X)\\
- intersects(X) (boolean, any overlap or touches)\\
- crosses(X) (boolean, the path passes across X, not along it)\\
- follows(X) (boolean, the vehicle's current path follows a line-string feature of type X, such as a road)\\
- opposes(X) (boolean, following the opposite direction)\\
- faces(X) (boolean, the vehicle is oriented towards X)\\

Available values for the current focal agent state are:\\
- speed: the current speed in km/h.\\
- acceleration: the current acceleration in (km/h)/s. Negative values indicate deceleration.\\
- yaw\_rate: the current absolute yaw rate in rad/s.\\
- is\_vehicle, is\_bus, is\_motorcyclist, is\_cyclist, is\_pedestrian.\\

HOW TO COMBINE PREDICATES:\\
You can write a conjunction/and as follows:
near\_stop\_sign if over(drivable\_area) and distance(vehicle) < 15.\\

You can write a disjunction/or by listing all options. There is NO direct "or" operator in Resin, so we list all options as separate rules:
on\_special\_lane if over(bike\_lane).
on\_special\_lane if over(bus\_lane).\\

You can also negate rules:\\
is\_no\_vehicle if not is\_vehicle.\\

Implications can be implemented via implication elimination, i.e. rewriting "a->b" as "not a or b". In Resin, this means writing two separate rules. For example, the implication "if over crossing then be\_slow" becomes:\\
landscape if not over(crossing).\\
landscape if speed < 5.0.\\

Close your generated rules with a "landscape if ..." clause. It is the logic result that captures the semantics of the user-provided natural language description.\\

SIMPLE EXAMPLES:\\
Input: Never drive on bus or bike lanes. Keep a safe distance from pedestrians.\\
Output:\\
\# We first define what it means to not be on a special lane:\\
on\_no\_special\_lane if not over(bus\_lane) and not over(bike\_lane).\\
\# We then define the landscape:\\
landscape if on\_no\_special\_lane and distance(pedestrian) > 2.\\

Input: One should be slow at crosswalks.\\
Output:\\
\# The implication is implemented as "not intersects(crossing) or speed < 5.0"\\
landscape if not intersects(crossing).\\
landscape if speed < 5.0.\\

CONSTRAINTS FOR AUTOMATION:\\
- Do NOT include any explanations, greetings, or conversational text.\\
- Output ONLY raw, syntactically correct Resin code.\\
- Do NOT wrap the output in Markdown blocks (no ```resin).\\
- Make sure to ALWAYS end with a "landscape if ..." clause that captures the semantics of the user-provided natural language description.\\
- Double-check that the output is syntactically correct Resin code before outputting.\\
- Double-check that the output captures the correct semantics of the user-provided natural language description.\\
\end{llmprompt}

\begin{llmprompt}{System Prompt for the Judging Agent}
You are an expert logic programming judge specializing in traffic rules in the new Resin language.\\

YOUR TASK:\\
Judge if the natural language descriptions of traffic rules are translated well into valid Resin code using world knowledge. It is key that the semantics are captured correctly.
Resin programs encode first-order logic rules into the final landscape predicate.
Comment lines start with a "\#".\\

AVAILABLE ATOMS AND PREDICATES:\\
The following atoms and predicates can be used in Resin. They will be automatically populated with real-time data when the Resin program is executed. New predicates can be derived as needed, but only based on the following atoms and existing rules.\\

Available map feature types are:\\
- Road: drivable\_area (union of all lanes).\\
- Road lanes: vehicle\_lane (open to all), bus\_lane, bike\_lane.\\
- Road lane center lines: vehicle\_lane\_centerline (open to all), bus\_lane\_centerline, bike\_lane\_centerline.\\
- Centerline of the lane of the focal agent: focal\_lane (>100 if unknown).\\
- Lane markings: solid\_line, dashed\_line.\\
- Special features: intersection, crossing.\\
- Moving traffic participants: vehicle, bus, cyclist, motorcyclist, pedestrian.\\

Available predicates, following Shapely semantics, for each feature type "X" are:\\
- over(X) (boolean)\\
- distance(X) (number, in meters)\\
- approaches(X) (boolean, the distance to X is decreasing)\\
- enters(X) (boolean, currently entering X)\\
- exits(X) (boolean, currently leaving X)\\
- intersects(X) (boolean, any overlap or touches)\\
- crosses(X) (boolean, the path passes across X, not along it)\\
- follows(X) (boolean, the vehicle's current path follows a line-string feature of type X, such as a road)\\
- opposes(X) (boolean, following the opposite direction)\\
- faces(X) (boolean, the vehicle is oriented towards X)\\

Available values for the current focal agent state are:\\
- speed: the current speed in km/h.\\
- acceleration: the current acceleration in (km/h)/s. Negative values indicate deceleration.\\
- yaw\_rate: the current absolute yaw rate in rad/s.\\
- is\_vehicle, is\_bus, is\_motorcyclist, is\_cyclist, is\_pedestrian.\\

COMBINING PREDICATES:\\
A conjunction/and is written as follows:\\
near\_stop\_sign if over(drivable\_area) and distance(vehicle) < 15.\\

Disjunction/or works by listing all options. There is NO direct "or" operator in Resin, so we list all options as separate rules:\\
on\_special\_lane if over(bike\_lane).\\
on\_special\_lane if over(bus\_lane).\\

One can also negate rules:\\
is\_no\_vehicle if not is\_vehicle.\\

Implications can be implemented via implication elimination, i.e. rewriting "a->b" as "not a or b". In Resin, this means writing two separate rules. For example, the implication "if over crossing then be\_slow" becomes:\\
landscape if not over(crossing).\\
landscape if speed < 5.0.\\

The rules always need to end with a "landscape if ..." clause. It is the logic result that captures the semantics of the user-provided natural language description.\\

JUDGING UNDER-SPECIFIED RULES:\\
Some natural-language rules cannot be fully captured by the available predicates (e.g. lane position, turn signals, and pedestrian intent are not directly observable). When this happens:\\
- Accept a reasonable, well-justified approximation as valid; do not demand exactness that the predicate set cannot express.\\
- Only reject when the program is semantically wrong: it inverts the direction of an implication, contradicts the natural-language rule, references undefined atoms, or collapses to a vacuous tautology (e.g. "landscape if X. landscape if not X.").\\
- Prefer accepting a rough-but-correct program over rejecting and forcing another iteration.\\

OUTPUT FORMAT:\\
Output valid JSON in the following format:
\begin{lstlisting}
{
  "verdict": true or false, // true if the Resin code correctly captures the semantics of the natural language description, false otherwise
  "critique": "a brief explanation of the judgment"
}
\end{lstlisting}

SIMPLE EXAMPLES:\\
Natural language: Never drive on bus or bike lanes. Keep a safe distance from pedestrians.\\
Resin code:\\
\# We first define what it means to not be on a special lane:\\
on\_no\_special\_lane if not over(bus\_lane) and not \\over(bike\_lane).\\
\# We then define the landscape:\\
landscape if on\_no\_special\_lane and distance(person) > 2.\\
Output:
\begin{lstlisting}
{
  "verdict": false,
  "critique": "The logic does not use the correct predicate for pedestrian distance. The correct feature type is 'pedestrian', not 'person'."
}
\end{lstlisting}

Natural language: One should be slow at crosswalks.\\
Resin code:\\
\# The implication is implemented as "not intersects(crossing) or speed < 5.0"\\
landscape if not intersects(crossing).
landscape if speed < 5.0.\\
Output:\\
\begin{lstlisting}
{
  "verdict": true,
  "critique": "The logic correctly captures the semantics of the natural language description."
}
\end{lstlisting}
\end{llmprompt}

\subsubsection{Traffic Rule 1 (Pedestrians)}
\begin{llmprompt}{User Prompt}
Pedestrians should not move faster than 5 km/h. They should only be on the drivable area if they are within 5 meters of a crossing. The final rule should be called pedestrian\_ok.
\end{llmprompt}

\begin{llmresponse}{Resin Logic Program}
\begin{lstlisting}
distance(crossing) <- source("/star_map/distance/crossing", Density).
over(drivable_area) <- source("/star_map/over/drivable_area", Probability).
speed <- source("/sensors/speed", Number).

# Define when a pedestrian is moving at acceptable speed
slow_pedestrian if not is_pedestrian.
slow_pedestrian if speed <= 5.

# Define when pedestrian is allowed on drivable area
# Implication: if pedestrian is over drivable_area, must be within 5m of crossing
pedestrian_on_road_ok if not is_pedestrian.
pedestrian_on_road_ok if not over(drivable_area).
pedestrian_on_road_ok if distance(crossing) < 5.

# Combine both into the final pedestrian_ok rule
pedestrian_ok if slow_pedestrian and pedestrian_on_road_ok.

landscape if pedestrian_ok.

landscape -> target("/landscape").\end{lstlisting}
\end{llmresponse}

\subsubsection{Traffic Rule 2 (Cyclists)}
\begin{llmprompt}{User Prompt}
Cyclists should not exceed 25 km/h. When a bike lane is within 5 meters, cyclists must use it unless they are at an intersection. The final rule should be called cyclist\_ok.
\end{llmprompt}

\begin{llmresponse}{Resin Logic Program}
\begin{lstlisting}
distance(bike_lane) <- source("/star_map/distance/bike_lane", Density).
over(bike_lane) <- source("/star_map/over/bike_lane", Probability).
over(intersection) <- source("/star_map/over/intersection", Probability).
speed <- source("/sensors/speed", Number).

# Cyclists must not exceed 25 km/h
cyclist_speed_ok if not is_cyclist.
cyclist_speed_ok if speed <= 25.

# If a bike lane is within 5 meters, cyclists must use it (be over it), unless at an intersection
cyclist_lane_ok if not is_cyclist.
cyclist_lane_ok if distance(bike_lane) >= 5.
cyclist_lane_ok if over(bike_lane).
cyclist_lane_ok if over(intersection).

# Final rule combining both
cyclist_ok if cyclist_speed_ok and cyclist_lane_ok.

landscape if cyclist_ok.

landscape -> target("/landscape").
\end{lstlisting}
\end{llmresponse}

\subsubsection{Traffic Rule 3 (Motorcyclists)}
\begin{llmprompt}{User Prompt}
Motorcyclists must stay on the drivable area and must not exceed 50 km/h. The final rule should be called motorcyclist\_ok.
\end{llmprompt}

\begin{llmresponse}{Resin Logic Program}
\begin{lstlisting}
over(drivable_area) <- source("/star_map/over/drivable_area", Probability).
speed <- source("/sensors/speed", Number).

# Rule for motorcyclists: must stay on drivable area and not exceed 50 km/h
# Implication: if is_motorcyclist then (over(drivable_area) and speed <= 50)
# Rewritten as: not is_motorcyclist or (over(drivable_area) and speed <= 50)

motorcyclist_ok if not is_motorcyclist.
motorcyclist_ok if over(drivable_area) and speed <= 50.

landscape if motorcyclist_ok.

landscape -> target("/landscape").
\end{lstlisting}
\end{llmresponse}

\subsubsection{Traffic Rule 4 (Buses)}
\begin{llmprompt}{User Prompt}
Buses must not turn with a yaw rate above 0.4 rad/s or accelerate faster than 5.0 (km/h)/s. When a bus lane is within 5 meters, buses must use it unless they are at an intersection. The final rule should be called bus\_ok.
\end{llmprompt}

\begin{llmresponse}{Resin Logic Program}
\begin{lstlisting}
distance(bus_lane) <- source("/star_map/distance/bus_lane", Density).
over(bus_lane) <- source("/star_map/over/bus_lane", Probability).
over(intersection) <- source("/star_map/over/intersection", Probability).
acceleration <- source("/sensors/acceleration", Number).
yaw_rate <- source("/sensors/yaw_rate", Number).

# Buses must not turn too sharply
no_sharp_turn if not is_bus.
no_sharp_turn if yaw_rate <= 0.4.

# Buses must not accelerate too fast
no_hard_accel if not is_bus.
no_hard_accel if acceleration <= 5.0.

# If a bus lane is within 5 meters and not at intersection, bus must use it
# Implication: (is_bus and distance(bus_lane) < 5 and not over(intersection)) -> over(bus_lane)
use_bus_lane if not is_bus.
use_bus_lane if distance(bus_lane) >= 5.
use_bus_lane if over(intersection).
use_bus_lane if over(bus_lane).

# Combine all bus rules
bus_ok if no_sharp_turn and no_hard_accel and use_bus_lane.

landscape if bus_ok.

landscape -> target("/landscape").
\end{lstlisting}
\end{llmresponse}

\subsubsection{Traffic Rule 5 (Lanes)}
\begin{llmprompt}{User Prompt}
Bike lanes are exclusively for cyclists and bus lanes are exclusively for buses. These restrictions do not apply at intersections. The final rule should be called lane\_type\_ok.
\end{llmprompt}

\begin{llmresponse}{Resin Logic Program}
\begin{lstlisting}
over(bike_lane) <- source("/star_map/over/bike_lane", Probability).
over(bus_lane) <- source("/star_map/over/bus_lane", Probability).
over(intersection) <- source("/star_map/over/intersection", Probability).

# Bike lanes are only for cyclists (unless at intersection)
bike_lane_ok if not over(bike_lane).
bike_lane_ok if is_cyclist.
bike_lane_ok if over(intersection).

# Bus lanes are only for buses (unless at intersection)
bus_lane_ok if not over(bus_lane).
bus_lane_ok if is_bus.
bus_lane_ok if over(intersection).

# Combine both conditions
lane_type_ok if bike_lane_ok and bus_lane_ok.

landscape if lane_type_ok.

landscape -> target("/landscape").
\end{lstlisting}
\end{llmresponse}

\subsubsection{Traffic Rule 6 (Vehicles)}
\begin{llmprompt}{User Prompt}
Vehicles should keep more than 5 meters from other vehicles. Do not accelerate toward a vehicle that is closer than 2 meters. The final rule should be called vehicle\_ok.
\end{llmprompt}

\begin{llmresponse}{Resin Logic Program}
\begin{lstlisting}
approaches(vehicle) <- source("/star_map/approaches/vehicle", Probability).
distance(vehicle) <- source("/star_map/distance/vehicle", Density).
acceleration <- source("/sensors/acceleration", Number).

# Define safe distance from other vehicles
safe_distance if distance(vehicle) > 5.

# Define no acceleration toward a very close vehicle
# Implication: if vehicle is closer than 2m and approaching, then do not accelerate
no_accel_toward_close if not approaches(vehicle).
no_accel_toward_close if distance(vehicle) >= 2.
no_accel_toward_close if acceleration <= 0.

# Final rule
vehicle_ok if safe_distance and no_accel_toward_close.

landscape if vehicle_ok.

landscape -> target("/landscape").
\end{lstlisting}
\end{llmresponse}

\subsubsection{Traffic Rule 7 (Safety Distances)}
\begin{llmprompt}{User Prompt}
Maintain at least 3 meters of distance from all vehicles, motorcyclists, and buses at all times. The final rule should be called min\_distance.
\end{llmprompt}

\begin{llmresponse}{Resin Logic Program}
\begin{lstlisting}
distance(bus) <- source("/star_map/distance/bus", Density).
distance(motorcyclist) <- source("/star_map/distance/motorcyclist", Density).
distance(vehicle) <- source("/star_map/distance/vehicle", Density).

# Define minimum distance from all required traffic participants
min_distance if distance(vehicle) > 3 and distance(motorcyclist) > 3 and distance(bus) > 3.

landscape if min_distance.

landscape -> target("/landscape").
\end{lstlisting}
\end{llmresponse}

\begin{table}[ht]
\centering
\caption{
    Per-model performance characteristics shown in~\Cref{fig:agent_capabilities}. 
    Higher is better for all metrics.
    \textit{Iter. Bu. Left} is the fraction of the refinement iteration budget not used by the agentic loop.
}
\label{tab:model_characteristics}
\setlength{\tabcolsep}{2pt}
\begin{tabular}{lcccccccc}
\toprule
   & \multicolumn{4}{c}{Pass Rate}
   & \multirow[b]{2}{*}{\makecell{Recov.\\Rate}}
   & \multirow[b]{2}{*}{\makecell{Iter.~Bu.\\Left}}
   & \multirow[b]{2}{*}{\makecell{Judge–Human\\Agreement}}
   & \multirow[b]{2}{*}{\makecell{Final Corr.\\Rate}} \\
  \cmidrule(lr){2-5}
   & Final & 1st Attempt & Syntax & Semantic & & & & \\
\midrule
  Qwen3 27B & 0.857 & 0.286 & 1.000 & 0.207 & 0.800 & 0.651 & 0.517 & 1.000 \\
  Claude Opus 4.7 & 1.000 & 1.000 & 1.000 & 1.000 & -- & 1.000 & 1.000 & 1.000 \\
  Gemini 3.1 Pro & 1.000 & 0.857 & 0.875 & 0.875 & 1.000 & 0.984 & 1.000 & 1.000 \\
  Gemma 4 31B & 1.000 & 0.857 & 1.000 & 0.875 & 1.000 & 0.984 & 0.750 & 0.857 \\
  GPT-5.5 & 1.000 & 1.000 & 1.000 & 1.000 & -- & 1.000 & 1.000 & 1.000 \\
\bottomrule
\end{tabular}
\end{table}

\newpage
\section{Machine Learning Setup}
\label{app:ml_setup}

This appendix describes the training, architecture, and inference details of the network that emits the per-scenario gating decision used by TraCS-gated fusion in \Cref{sec:experiments}.
The fusion mechanism itself (log-linear pooling with weight $w$) is described in \Cref{sec:fusion} and held fixed across all experiments.

\subsection{Dataset and splits}
\label{app:data}

We evaluate on the Argoverse~2 Motion Forecasting Dataset (AV2)~\citep{Argoverse2}. 
All three backbones (\textsc{DeMo}, \textsc{Polaris}, \textsc{QCNet}) are run on the AV2 \emph{val} split ($n{=}24{,}988$), which we use to \emph{train} the
trust classifier, and the AV2 \emph{test} split ($n{=}24{,}984$), on which we report all numbers in \Cref{sec:experiments}. 
The val split is further partitioned $90/10$ into a training fold ($n{=}22{,}490$) and an internal validation fold ($n{=}2{,}498$) used for early stopping, model selection, and gating-threshold selection. 
No backbone, TraCS, or fusion parameters are tuned on the AV2 test split.

\subsection{Inputs and features}
\label{app:features}

The gating network consumes a flat per-scenario feature vector $\mathbf{x}\in\mathbb{R}^{D}$ ($D{=}299$) that concatenates two groups: a small set of hand-crafted scalars summarising the scoring statistics of the scenario, and the backbone's own scene-level latent embedding. 
\Cref{tab:features} lists every input dimension together with its definition. 
Feature number and order is fixed across scenarios, which lets us split the input by index at the model boundary in \Cref{app:arch}.

\begin{table}[t]
\centering
\caption{Per-scenario input features to the gating network. All hand-crafted
scalars are derived from the backbone score vector
$\mathbf{p}^{\text{bb}}\!\in\!\Delta^{6}$, the TraCS score vector
$\mathbf{p}^{\text{tr}}\!\in\!\Delta^{6}$ over the six candidate
trajectories, the predicted endpoint coordinates, and the focal-agent class
metadata. The backbone latent embedding is obtained once per backbone from
its penultimate-layer representation and joined to the scenario by id.}
\label{tab:features}
\small
\renewcommand{\arraystretch}{1.15}
\begin{tabular}{@{}lrl@{}}
\toprule
Feature group & Dim & Description \\
\midrule
\multicolumn{3}{@{}l}{\textit{Hand-crafted scoring statistics}} \\
Raw backbone scores              &   6 & $p^{\text{bb}}_k$ for each candidate $k\!=\!1,\ldots,6$. \\
Sorted backbone scores           &   6 & Backbone scores sorted in descending order. \\
Backbone entropy                 &   1 & Shannon entropy $H(\mathbf{p}^{\text{bb}})$ in bits. \\
Raw TraCS scores                 &   6 & $p^{\text{tr}}_k$ for each candidate $k$. \\
Sorted TraCS scores              &   6 & TraCS scores sorted in descending order. \\
TraCS entropy                    &   1 & Shannon entropy $H(\mathbf{p}^{\text{tr}})$ in bits. \\
Spearman correlation             &   1 & Rank correlation between $\mathbf{p}^{\text{bb}}$ and $\mathbf{p}^{\text{tr}}$. \\
KL divergences                   &   2 & $D_{\mathrm{KL}}(\mathbf{p}^{\text{bb}}\!\parallel\!\mathbf{p}^{\text{tr}})$ and reverse. \\
Entropy difference               &   1 & $H(\mathbf{p}^{\text{tr}}) - H(\mathbf{p}^{\text{bb}})$. \\
Top-2 score margins              &   2 & Top-1 minus top-2 score, for backbone and for TraCS. \\
Cross-classifier top scores      &   2 & TraCS score at backbone's argmax and vice versa. \\
Cross-classifier top ranks       &   2 & Rank under each classifier of the other's argmax. \\
Predicted endpoint displacement  &   2 & Mean and standard deviation of endpoint $\lVert\cdot\rVert_2$ across candidates. \\
Focal-agent class                &   5 & One-hot over \{vehicle, bus, cyclist, motorcyclist, pedestrian\}. \\
\addlinespace[2pt]
& \textit{43}              & \textit{Subtotal} \\
\midrule
\multicolumn{3}{@{}l}{\textit{Backbone latent embedding from penultimate-layer}} \\
Focal-agent latent               & 128   & Focal agent embedding. \\
Scene-averaged latent            & 128   & Average non-focal agent embedding. \\
Scene-averaged map embedding     & 128 & Average map embedding (only QCNet). \\
\addlinespace[2pt]
& \textit{256}              & \textit{Subtotal (384 for QCNet)} \\
\midrule
\textbf{Total input dimension}                     & $\mathbf{D = 299}$ & \\
\bottomrule
\end{tabular}
\end{table}

\subsection{Architecture}
\label{app:arch}

The trust network is a small two-branch multi-layer perceptron that emits a single scalar trust score $\hat{w}\in[0,1]$:
\begin{align}
    \hat{w} \;=\; \sigma\big(\,\textsc{MLP}_{\text{trust}}\big(
   [\,\mathbf{x}^{\text{feat}}\,;\;\textsc{MLP}_{\text{emb}}(\mathbf{x}^{\text{emb}})\,]
\big)\,\big).
\end{align}
The embedding branch projects the $256$-d backbone latent through a heavily-regularised bottleneck to keep it from dominating the ${\approx}40$ hand-crafted scalars:
\begin{align}
  \textsc{MLP}_{\text{emb}} \;:\;
   \mathbb{R}^{256} \xrightarrow{\text{BN}}
   \xrightarrow{\text{Linear}_{256\to 32}}
   \xrightarrow{\text{ReLU}}
   \xrightarrow{\text{Dropout}_{0.5}} \mathbb{R}^{32}.
\end{align}
The trust network then operates on the concatenated $n_{\text{feat}}\!+\!32 = d'$ features:
\begin{align}
  \textsc{MLP}_{\text{trust}} \;:\;
   \mathbb{R}^{d'}
   \xrightarrow{\text{BN}}
   \xrightarrow{\text{Linear}_{\to 128}}
   \xrightarrow{\text{ReLU}, \text{Dropout}_{0.3}}
   \xrightarrow{\text{Linear}_{128\to 64}}
   \xrightarrow{\text{ReLU}, \text{Dropout}_{0.2}}
   \xrightarrow{\text{Linear}_{64\to 1}}
   \xrightarrow{\sigma(\cdot)}
   [0, 1].
\end{align}
The input BatchNorm normalises the wildly different scales of the hand-crafted features (entropy in nats, KL divergence, raw probabilities, etc.).
The heavy dropout on the embedding branch ($p{=}0.5$) is essential -- raw $256$-d latents fed directly to the trunk overfit the trust-classification task in under ten epochs; the projection plus dropout collapses the train/validation gap to a stable margin. 
Total parameter count is about $50\text{k}$.

\subsection{Supervision target}
\label{app:supervision}

For each scenario in the training fold we form a per-scenario regression target $w^{\star}\!\in\![0,1]$ by grid-searching the scalar fusion weight that minimises the expected average-displacement error (ADE) under the fused distribution:
\begin{align}
  w^{\star}
   \;=\; \argmin_{w \in [0,1]}\;
         \sum_{k=1}^{6} p^{\text{fused}}_k(w)\,\cdot\,\text{ADE}_k,
\end{align}
evaluated on a uniform grid of $1000$ values of $w$. The choice of expected ADE (rather than the brittle winner-take-all bMinADE$_1$) gives a smooth, unimodal optimisation landscape per scenario, while the empirical distribution of $w^{\star}$ across scenarios remains strongly bimodal at $0$ and $1$, i.e., the supervisor itself is implicitly making a hard trust decision. 
This bimodality of the targets motivates both our loss choice and our inference policy below.

\subsection{Training}
\label{app:training}

The gating network is trained with Binary Cross-Entropy (BCE) on the soft target
$w^{\star}\in[0,1]$:
\begin{align}
  \mathcal{L}(\hat w,w^{\star})
   \;=\;
   -\,w^{\star}\log \hat{w}\;-\;(1-w^{\star})\log(1-\hat{w}).
\end{align}
We use Adam~\citep{DBLP:journals/corr/KingmaB14adam} with learning rate $10^{-3}$, weight decay $10^{-4}$, and mini-batches of size $256$. 
Learning rate is decayed by $0.5\!\times$ on plateaus of the validation loss (\texttt{ReduceLROnPlateau}, patience $15$ epochs, floor $10^{-6}$).
Training runs for up to $500$ epochs with early stopping on validation loss (patience $30$, $\min\Delta = 10^{-5}$); on every backbone, the training terminates well before the cap, and we restore the parameters at the best validation epoch.

\begin{figure}[t]
  \centering
  \includegraphics[width=\textwidth]{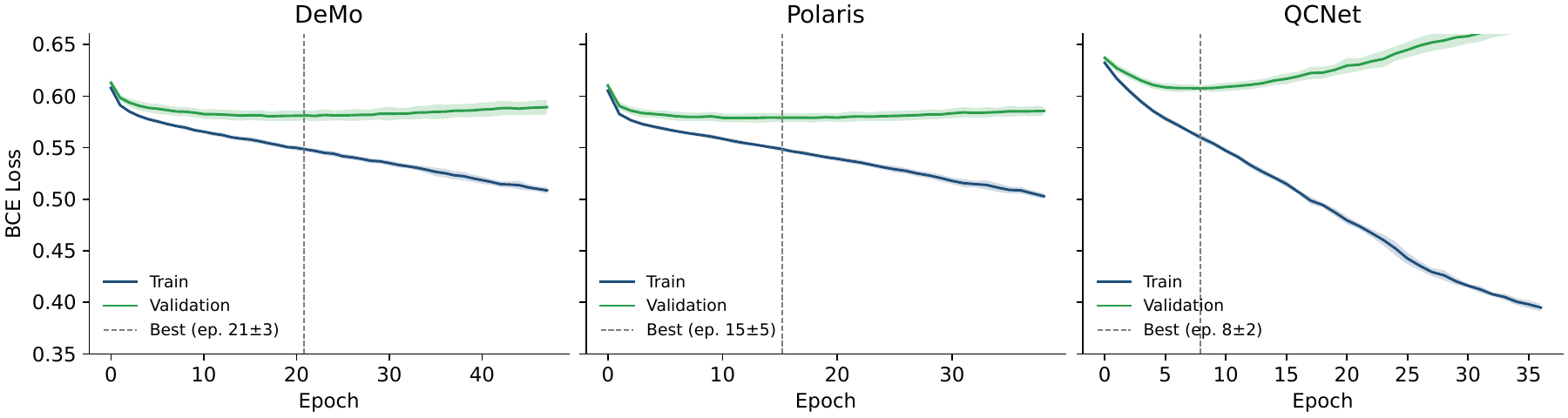}
  \caption{\textbf{Training curves.}
    Training and validation BCE loss for the gating network trained against each of the three backbones;
    the dashed vertical line marks the best-validation epoch at which model weights are restored. 
    All three runs converge to a stable plateau within ${\approx}50$ epochs.}
  \label{fig:learning_curves}
\end{figure}

\paragraph{Learning curves.}
\Cref{fig:learning_curves} shows the training and validation loss trajectories for each backbone. 
All three converge to a stable plateau within ${\approx}50$ epochs; the small persistent train/validation gap is bounded by the embedding-branch regularisation (\Cref{app:arch}).

\subsection{Inference and gating}
\label{app:gating}

At inference, the gating network produces a real-valued trust score $\hat{w}\in[0,1]$.
We then apply a binary gate at a fixed threshold $\theta\!\in\![0,1]$:
\begin{align}
  w_{\text{gate}}(\hat{w};\theta)
   \;=\;
   \begin{cases}
     1, & \hat{w} \ge \theta,\\
     0, & \hat{w} < \theta,
   \end{cases}
\end{align}
and pass $w_{\text{gate}}$ into the log-linear fusion of \Cref{sec:experiments}. 
The binary policy reflects two empirical findings: (i) the supervisor target $w^{\star}$ is strongly bimodal at the endpoints, so committing to $\{0,1\}$ matches the grid-search optimum on most scenarios; (ii) on a held-out validation fold the binary policy dominates the continuous policy ($w_{\text{cont}}=\hat{w}\!\cdot\!\indicator[\hat{w}\!\ge\!\theta]$) across all thresholds we evaluated.

\begin{figure*}[t]
  \centering
  \includegraphics[width=\textwidth]{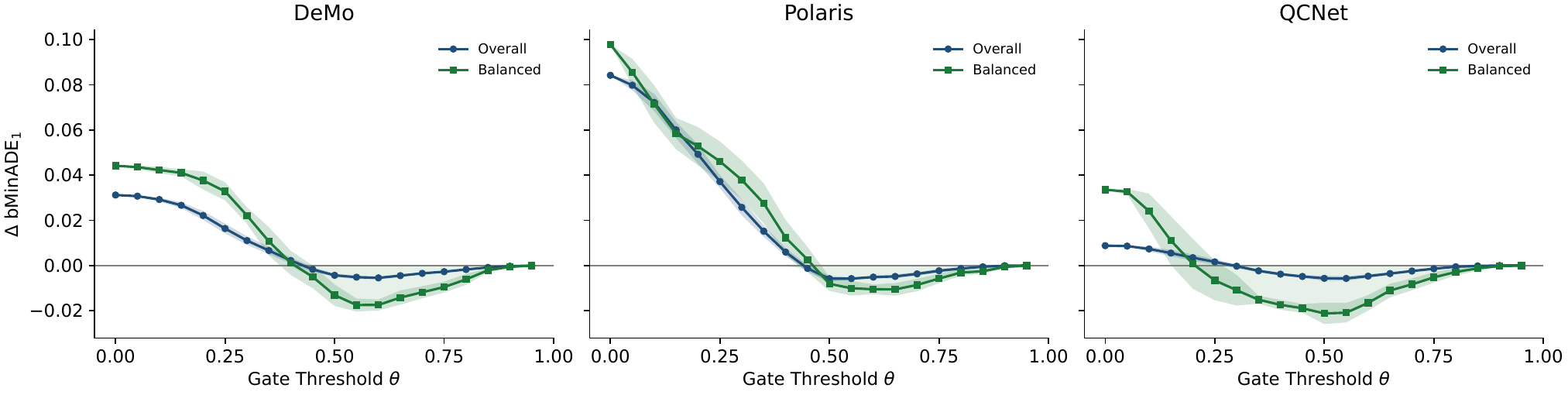}
  \caption{\textbf{Gating-threshold sweep on the test split.} 
    Change in overall and class-balanced Brier-minADE$_1$ as a function of the binary gate threshold $\theta$ for each of the three backbones. 
    Ungated fusion ($\theta\!=\!0$) is harmful on average; gating drives
    the change into the green region across a broad plateau, indicating that
    the selected operating point is robust to the precise choice of
    $\theta$.}
  \label{fig:gating_sweep}
\end{figure*}

\paragraph{Threshold selection.}
The threshold $\theta$ is selected \emph{per backbone} on the held-out validation data fold by sweeping $\theta\!\in\!\{0.0,0.05,\ldots,0.95\}$ and choosing the value that minimizes the validation bMinADE$_1$.
\Cref{fig:gating_sweep} shows the test-set version of the same sweep for each backbone averaged over 10 seeds. 
The curves share a consistent structure: at $\theta\!=\!0$ (ungated) fusion is \emph{harmful} on average, gating drives the change negative, and a broad plateau of near-optimal thresholds exists in the $\theta\!\in\![0.5,0.7]$ range across all three backbones, indicating the operating point is robust rather than a knife-edge tune.

\paragraph{Reporting.}
All numbers reported in \Cref{sec:experiments} use the validation-selected $\theta$ applied as a fixed scalar at test time. 
No per-scenario or per-class threshold tuning is performed; the same gate operates uniformly across all $24{,}984$ AV2 test scenarios.

\end{document}